\documentclass[twoside]{article}

\usepackage{aistats2021}
\usepackage{graphicx}
\usepackage{multicol}
\usepackage{multirow}
\usepackage{rotating}
\usepackage{enumitem}
\usepackage{booktabs}
\usepackage{subfig}
\usepackage{amsfonts}
\usepackage{stfloats}

\usepackage{xcolor}
\usepackage{pifont}% http://ctan.org/pkg/pifont
\newcommand{\cmark}{\ding{51}}%
\newcommand{\xmark}{\color{lightgray}\ding{55}}%
\newcommand{\dash}{\color{lightgray}\textbf{--}}%

\renewcommand\vec{\boldsymbol}
\usepackage{float}
\usepackage{natbib}
\usepackage{amsfonts}

\DeclareMathOperator*{\argminB}{argmin}
\begin{document}

\twocolumn[

\aistatstitle{Model-Attentive Ensemble Learning for Sequence Modeling}

\aistatsauthor{ Victor D. Bourgin$^*$ \And Ioana Bica \And  Mihaela van der Schaar }

\aistatsaddress{ University of Cambridge \And  University of Oxford\\The Alan Turing Institute  \And University of Cambridge, UCLA\\The Alan Turing Institute } ]

\begin{abstract}
\vspace{-0.2cm}
Medical time-series datasets have unique characteristics that make prediction tasks challenging. Most notably, patient trajectories often contain longitudinal variations in their input-output relationships, generally referred to as temporal conditional shift. Designing sequence models capable of adapting to such time-varying distributions remains a prevailing problem. To address this we present Model-Attentive Ensemble learning for Sequence modeling (MAES). MAES is a mixture of time-series experts which leverages an attention-based gating mechanism to specialize the experts on different sequence dynamics and adaptively weight their predictions. We demonstrate that MAES significantly out-performs popular sequence models on datasets subject to temporal shift.
\end{abstract}

\vspace{-0.6cm}
\section{Introduction}
\vspace{-0.2cm}
With the advent of electronic health records, time-series (TS) modeling has become an invaluable tool for clinical decision support (\cite{healthcare_mimic_mortality_prediction, healthcare_mortality_prediction, deeplearning_healthcare}). However, one pervasive problem common to many sequential medical datasets is Temporal Conditional Shift (TCS) (\cite{HyperLSTM, SMS-DKL, risk_evolution_TCS}). TCS corresponds to longitudinal variations in the input-output distribution; for example, the relationship between a patient's vitals and their risk factors may change over the course of their hospital stay as their health starts deteriorating.

Popular sequence models such as Long Short-Term Memory (LSTM) struggle to adapt to TCS due to their complete sharing of parameters (\cite{HyperLSTM}). The latter often causes models to capture \textit{average} trends across sequences, disregarding the evolving relationships in the data -- \cite{HyperLSTM} termed this \textit{temporal bias}. Various techniques have been developed to tackle this problem, such as altering the LSTM architecture to relax this parameter-sharing, or combining the predictions of multiple base models with different hypothesis spaces (\cite{HyperLSTM, stackingLSTM, adaptive_weighting_ensemble_LSTM, ADE}).

However, LSTM adaptations are often complex and most work on Ensemble Learning (EL) uses independently trained base learners (\cite{stackingLSTM, adaptive_weighting_ensemble_LSTM, ADE}). Independently trained base learners suffer from the same limitations as `classic' single-model techniques; they tend to model average dynamics in the data, thereby leading to high temporal biases (\cite{HyperLSTM}). An ensemble would benefit more from the combination of specialized models (\textit{experts}), each capturing different local conditional relationships in sequences. Additionally, most EL methods learn a fixed set of aggregation weights for the base learners (\cite{Stacking_original, stackingLSTM, EL_review}). A desirable characteristic would be for the ensemble’s combination weights to depend on both the prediction step and the history of patient features, as different patients are likely to exhibit different variations in their feature-to-outcome distribution at different times.

To this end we present Model-Attentive Ensemble learning for Sequence modeling (MAES). MAES is a novel EL method for temporal data, consisting of multiple sequence models whose predictions are combined using an attention-based gating mechanism. To address the problem of TCS, the base models specialize on different sequence dynamics through a gating architecture and training procedure inspired from Mixtures-of-Experts (ME) (\cite{ME_original}), and the gate adaptively combines their predictions according to the patient’s trajectory. Through a set of experiments on synthetic datasets with simulated TCS, we demonstrate MAES' superior performance compared to baseline single-model techniques and ensembles, and provide insights into the sources of improvement.

\section{Problem Formulation}
\vspace{-0.4cm}
\subsection{On-line sequence prediction}
\vspace{-0.3cm}
Let $\mathcal{D} = \{\vec{s}_n,\vec{x}_{n,1:T} \}_{n=1}^N$ represent a clinical dataset with $N$ patients, where $\vec{s}_n$ and $\vec{x}_{n,1:T}$ denote patient $n$’s static and temporal variables respectively. For simplicity, throughout this work we assume that all sequences are uniformly sampled and have equal length $T$. Although the techniques developed here are applicable to many prediction tasks (and variable-length sequences), for illustration we focus on the on-line classification task, where a categorical label $\vec{y}_{n,{t}}$ is issued at every time step $t$ given $H_{n,1:t} = (\vec{s}_n,\vec{x}_{n,1:t})$. In a clinical setting, this could correspond to predicting a patient's need for a ventilator in the following hour.

\vspace{-0.3cm}
\subsection{Temporal conditional shift}
\vspace{-0.2cm}
In most TS models, temporal relationships are assumed to be governed by a constant function:

\vspace{-0.5cm}
\begin{equation}
    \vec{y}_{n,t} = f(H_{n,1:t})
\end{equation}
\vspace{-0.7cm}

Parametric models such as LSTMs approximate the true conditional distribution through a fixed set of parameters $\theta$ and hyperparameters $\mu$:

\vspace{-0.5cm}
\begin{equation}
    p(\vec{y}_{n,t}|H_{n,1:t}) = \hat{f}(H_{n,1:t};\theta,\mu)
\label{eq:predictive_parametric_global}
\end{equation}
\vspace{-0.7cm}

Given some pre-defined hyperparameters $\mu$, the model parameters $\theta$ are generally found through Maximum Likelihood Estimation (MLE) on the training data $\mathcal{D}$. For categorical labels, this generally corresponds to minimizing the Cross-Entropy (CE) loss over the sequences:

\thinmuskip=0.4mu
\thickmuskip=0.4mu
\scriptspace=0pt

\vspace{-0.9cm}
\begin{equation}
    \theta^* = \text{argmin}_{\theta\in\Theta} -\sum_{n=1}^N\sum_{t=1}^{T}\log\prod_{k=1}^K\hat{f}^k({H}_{n,1:t};\theta,\mu)^{\vec{y}_{n,t}^k}
    \label{eq:CE_loss}
\end{equation}
\vspace{-0.6cm}

\thinmuskip=3mu
\thickmuskip=5mu
\scriptspace=0.5pt

Where $K$ is the number of classes and $\hat{f}^k(\cdot)$ and $\vec{y}_{n,t}^k$ denote the $k^{th}$ element of the model prediction and target respectively.

With TCS, however, the input-output distribution varies along the sequence: 

\vspace{-0.5cm}
\begin{equation}
    \vec{y}_{n,t} = f(H_{n,1:t},t)
\label{eq:inputoutput_relationship_TCS}
\end{equation}
\vspace{-0.7cm}

For time-varying relationships governed by Equation \ref{eq:inputoutput_relationship_TCS}, training a sequence model according to Equation \ref{eq:CE_loss} may lead to temporal biases. In the case of LSTMs for example, the parameters are shared across time steps such that the entire time-dependence in the distribution must be indirectly incorporated in the hidden state and cell state (\cite{HyperLSTM}; cf. supplementary material). Capturing temporal variations in the distribution has therefore proven difficult for LSTMs, particularly in the presence of limited training data (a ubiquitous concern in healthcare), and models tend capture average trends instead (\cite{HyperLSTM}). In the following section, we present adaptation techniques that have been developed to tackle this.

\section{Related Work}
\label{section:related_work}
\vspace{-0.3cm}

\textbf{Single-Model Techniques}. TCS adaptation can be achieved by learning time-dependent parameters $\theta_t$:

\vspace{-0.5cm}
\begin{equation}
    p(\vec{y}_{n,t}|H_{n,1:t}) = \hat{f}(H_{n,1:t};\theta_t,\mu)
\end{equation}
\vspace{-0.7cm}

\cite{HyperLSTM} designed a `mixLSTM' with relaxed parameter sharing. Specifically, different LSTM cells and parameters are learned on arbitrarily partitioned subsets of the sequences and combined through time-dependent mixing coefficients. \cite{hypernetworks} used hypernetworks instead to modulate LSTM weights along sequences.

However, these methods require complex architectural adaptations to the `classic' LSTM, and the arbitrary partitioning of the time series may be sub-optimal. To circumvent these issues, \cite{SMS-DKL} developed Step-wise Model Selection via Deep Kernel Learning (SMS-DKL), where a different `classic' LSTM is selected to issue a prediction at every time step:

\vspace{-0.5cm}
\begin{equation}
    p(\vec{y}_{n,t}|H_{n,1:t}) = \hat{f}(H_{n,1:t};\theta^*,\mu^*_t)
\end{equation}
\vspace{-0.7cm}

Where $\mu^{*}_t$ is the optimal set of LSTM hyperparameters at time step $t$, and $\theta^*$ are the corresponding model parameters learned through MLE. The authors developed a novel Bayesian Optimization (BO) procedure to identify optimal LSTM hyperparameters for each prediction step. However, each model is still trained to issue predictions over the whole sequence (Equation \ref{eq:CE_loss}), thus still potentially modeling the average dynamics and limiting the performance improvements of step-wise selection with TCS.

\textbf{Ensemble Learning}. Other studies employed an EL approach to TCS adaptation. The rationale for using EL is that heterogeneous models have different hypothesis spaces, such that each one may capture a different aspect of the distribution. Combining their predictions with time-dependent
aggregation weights may therefore enable the ensemble to adapt to time-varying dynamics:

\vspace{-0.9cm}
\begin{equation}
    p(\vec{y}_{n,t}|H_{n,1:t}) = \sum_{m=1}^M w_{m,t} \hat{f}(H_{n,1:t};\theta_m,\mu_m)
\label{eq:simple_EL}
\end{equation}
\vspace{-0.7cm}

Where $M$ is the number of base models and $\{\{w_{m,t}\}_{t=1}^T\}_{m=1}^M$ is a set of aggregation weights.

Ensemble learning models differ mainly in their selection and training methods for the base learners and the aggregation weights. \cite{adaptive_weighting_ensemble_LSTM} trained homogeneous LSTMs on different sequence lengths and learned different combination weights at each time step according to the base learners' prediction errors on validation data (we call this work `Adaptive LSTM Weighting'). Similarly, \cite{stackingLSTM} showed empirically that a stacking ensemble of LSTMs can adapt better to variations in the data than a single LSTM (`Stacking LSTM'). In their work, diversity was introduced through a hybrid method: each LSTM was trained on a different sequence length with different hyperparameter values.

\vspace{-0.05cm}
However, in all cases, the combination weights are independent of the specific characteristics of the data. \cite{ADE} designed an Arbitrated Dynamic Ensemble (ADE). ADE is an adaptive, instance-dependent ensemble model that combines the base learners’ outputs as a function of their prediction error on previous time steps. After training $M$ simple base learners independently, $M$ separate meta-models are trained on a validation set to model the errors of their base learner counterparts. The meta-models take as input the instances directly and their outputs are used to weight the base learner predictions in the final ensemble.

\vspace{-0.05cm}
The principal specificity of MAES compared to all methods investigated above is that in the latter, all base learners were trained independently and separately from the combination weights. Although these weights were designed to vary along sequences, the high temporal biases in the base models' predictions resulting from their independent training limit the ability of the ensembles to adapt to TCS. Instead, MAES jointly specializes experts on different conditional distributions and adaptively weights their contributions along sequences. Table \ref{table:previous_work} summarizes the differences between our model and previous work.

\setlength\textfloatsep{10pt}

\setlength{\tabcolsep}{2pt}
\renewcommand{\arraystretch}{1}
\begin{table*}[htbp]
\caption{Comparative table with previous work aimed at adapting to sequence dynamics. $^\dagger$A different model's predictions are selected at each time step. mixLSTM: \cite{HyperLSTM}; SMS-DKL: \cite{SMS-DKL}; Stacking LSTM: \cite{stackingLSTM}; Adaptive Weighting LSTM: \cite{adaptive_weighting_ensemble_LSTM}; ADE: \cite{ADE}. }
\centering
\vspace{-0.2cm}
\begin{tabular}{ccc@{\hskip 2.5em}cccc}
\toprule
\multirow{2}{*}{} & \multicolumn{2}{l}{\qquad\textbf{Single-Model}} & \multicolumn{4}{c}{\textbf{Ensemble}} \\ \cline{2-7} 
 & \begin{tabular}[c]{@{}c@{}}mixLSTM\\[0.0cm] \end{tabular} & \begin{tabular}[c]{@{}c@{}}SMS-DKL\\[0.0cm] \end{tabular} & \begin{tabular}[c]{@{}c@{}}Stacking \\[0.0cm]LSTM \end{tabular} & \begin{tabular}[c]{@{}c@{}}Adaptive \\[0.0cm]Weighting LSTM \end{tabular} & \begin{tabular}[c]{@{}c@{}}ADE\\[0.0cm] \end{tabular} & \textbf{MAES} \\ \midrule
\begin{tabular}[c]{@{}c@{}}Time-Adaptive Model \end{tabular} & \cmark & \cmark & \cmark & \cmark & \cmark & \cmark \\[0.0cm]
\begin{tabular}[c]{@{}c@{}}Time-Dependent Combination\end{tabular} & \dash & \cmark$^\dagger$ & \cmark & \cmark & \cmark & \cmark \\[0.0cm]
\begin{tabular}[c]{@{}c@{}}Feature-Dependent Combination\end{tabular} & \dash & \dash & \xmark & \xmark & \cmark & \cmark \\[0.0cm]
\begin{tabular}[c]{@{}c@{}}Base Model Specialization\end{tabular} & \dash & \dash & \xmark & \xmark & \xmark & \cmark \\
\bottomrule
\end{tabular}%
\label{table:previous_work}
\vspace{-0.35cm}
\end{table*}

\vspace{-0.3cm}
\section{Model-Attentive Ensemble learning for Sequence modeling (MAES)}
\vspace{-0.2cm}
MAES is a mixture of sequence model experts whose predictions are combined through an attention-based gating mechanism. We derive the governing equation for MAES and describe its key components and training procedure.

\vspace{-0.3cm}
\subsection{Mixture of Experts}
\vspace{-0.2cm}
In order to design an EL method capable of adapting to time-varying conditional distributions while incorporating feature dependence, we set the following requirements for MAES:

\vspace{-0.4cm}
\begin{enumerate}[start=1,label={\bfseries R\arabic*:}, wide, labelwidth=!, labelindent=0pt]
    \item \textbf{Base Model Specialization}. The base models must be trained jointly using a training method that encourages specialization on different input-output distributions.
    \vspace{-0.1cm}
    \item \textbf{Adaptation to Instance Dynamics}. The base learners' relative contribution to the ensemble prediction must depend on their predictive ability on both the instance features and the prediction step.
\end{enumerate}
\vspace{-0.3cm}

Equation \ref{eq:simple_EL} can be modified to allow the model contributions to depend on the instance features and time:

\thinmuskip=0.1mu
\thickmuskip=0.1mu
\scriptspace=0pt

\vspace{-0.7cm}
\begin{equation}
 p(\vec{y}_{n,t}|H_{n,1:t})= \sum_{m=1}^M w_m\left({H}_{n,1:t};\theta_w\right) \hat{f}({H}_{n,1:t};\theta_m,\mu_m)
\label{eq:simple_ME}
\end{equation}
\vspace{-0.5cm}

\thinmuskip=3mu
\thickmuskip=5mu
\scriptspace=0.5pt

Where $w_m(\cdot)$ are instance-dependent aggregations weights parametrized by $\theta_w$. From a Bayesian perspective, these weights approximate the model posterior $p(\theta_m,\mu_m|{H}_{n,1:t})$ (cf. supplementary material). Equation \ref{eq:simple_ME} closely resembles ME. In ME, a parametric \textit{gate} controls each base model’s contribution to the ensemble, with gating weights that depend on the input features. The ME training procedure jointly trains the gate and the base models, encouraging the ensemble to learn a soft partitioning of the input-output space which can be modeled using a few specialized base learners -- these specialized models are called \textit{experts} (\cite{ME_original}). Here, the design of the gate is not straightforward: the input is a sequence, and we want the weights to incorporate dependence on both the instance features and time.

\vspace{-0.3cm}
\subsection{Attention Gating}
\vspace{-0.2cm}
In this section, we justify the use of an attention model for the gate -- intuitively, we are interested in evaluating how well each model ``aligns'' with a given input sequence at a given time when weighting their prediction. We first propose a probabilistic motivation for the attention-based gate before describing its core design considerations.

\vspace{-0.3cm}
\subsubsection{Probabilistic Perspective}
\vspace{-0.2cm}
One significant strength of attention which has motivated its use for a wide variety of tasks is that it is theoretically Turing complete \citep{attention_turingcomplete}. While many different types of attention models have been introduced to achieve state-of-the-art results in specific applications in Neural Machine Translation (NMT) and computer vision (\cite{transformer, self-attention_image_recognition, CDSA_selfattention_imputation}), \citet{show_attend_tell} described two broad classes of attention mechanisms with different probabilistic interpretations: hard (stochastic) and soft (deterministic) attention. We describe and extend these concepts to model selection and ensemble learning.

\vspace{-0.1cm}
\textbf{Hard (stochastic) Attention}. Let us denote $\vec{\eta}_{t}^n$ the model selection random variable, which is an $M$-dimensional latent variable indicating which model is used to issue a prediction for ${H}_{n,1:t}$: $\eta_{t,m}^n=1$ if model $(\mu_m,\theta_m)$ is used for the prediction, and $\eta_{t,m}^n=0$ otherwise. With stochastic attention, a categorical distribution is assigned to $\vec{\eta}_{t}^n$, parametrized by attention weights $\alpha_{t,m}^n$. The attention weights then represent the probability that model $m$ (the key / value) is the right one to focus on when issuing a prediction for the input sequence ${H}_{n,1:t}$ (the query): $\alpha_{t,m}^n=p(\eta_{t,m}^n =1|{H}_{n,1:t},\mathcal{D})=p(\theta=\theta_m,\mu=\mu_m|{H}_{n,1:t})$. The predictive distribution $p(\vec{y}_{n,t}|{H}_{n,1:t})$ can then be viewed as a random variable itself:

\vspace{-0.5cm}
\begin{equation}
    p(\vec{y}_{n,t}|{H}_{n,1:t})=\sum_{m=1}^{M}\eta_{t,m}^n\hat{f}(H_{n,1:t};\theta_m,\mu_m)
\end{equation}
\vspace{-0.4cm}

Here stochastic attention corresponds to model selection: the latent variable $\vec{\eta}_{t}^n$ is a one-hot encoding drawn from a categorical distribution, which selects a single model for issuing a prediction for sequence $n$ at time $t$.

\vspace{-0.1cm}
\textbf{Soft (deterministic) attention}. Under a soft attention interpretation, we compute the expectation of the base learners' predictive distribution over the latent distribution $p(\vec{\eta}_{t}^n|{H}_{n,1:t})$:

\thinmuskip=0.2mu
\thickmuskip=0.2mu
\scriptspace=0.5pt

\vspace{-0.6cm}
\begin{align}
    p(\vec{y}_{n,t}|{H}_{n,1:t})&=\sum_{m=1}^{M}p(\eta_{t,m}^n =1|{H}_{n,1:t})\hat{f}(H_{n,1:t};\theta_m,\mu_m)\nonumber
    \\
    &=\sum_{m=1}^{M}\alpha_{t,m}^n\hat{f}(H_{n,1:t};\theta_m,\mu_m)
    \label{eq:attention_predictive}
\end{align}
\vspace{-0.5cm}

\thinmuskip=3mu
\thickmuskip=5mu
\scriptspace=0.5pt

With a soft attention mechanism we recover Equation \ref{eq:simple_ME} for finite instance-dependent ensembles with $\alpha_{t,m}^n=w_m\left(H_{n,1:t};\theta_w\right)$, since $\alpha_{t,m}^n=p(\theta=\theta_m,\mu=\mu_m|{H}_{n,1:t})$ (the nature of the parameters $\theta_w$ depend on the attention mechanism architecture). In essence, attention represents the probability that model $m$ is the right model to focus on for the prediction step. We highlight that this corresponds to a different interpretation of attention than that traditionally employed in NMT: in NMT, attention is used to create a context vector from multiple source encodings whereas here, attention is computed at the model level for aggregating base learner predictions in an ensemble.

\vspace{-0.3cm}
\subsubsection{Attention Gate Design}
\vspace{-0.2cm}
With an attention-based gate, the predictive distribution is computed according to Equation \ref{eq:attention_predictive}. We now describe our attention gating mechanism designed to emulate $p(\theta_m,\mu_m|{H}_{n,1:t})$. 

\vspace{-0.1cm}
We need finite-length representations of both variables involved in the model posterior: the sub-sequence ${H}_{n,1:t}$ and the base model $m$. Let the context $\vec{c}_{n,t}$ be an encoding of instance $n$'s features at time $t$, i.e. $\vec{c}_{n,t}=f_c(H_{n,1:t})$ for a mapping function $f_c$. We represent each expert $m$'s predictive capability on different temporal contexts through a vector $\vec{u}_m$, which we describe in further detail below. We can then compute attention weights from the alignment between each expert encoding and the context:

\vspace{-0.5cm}
\begin{equation}
    \alpha_{t,m}^n = \frac{exp\left(f_{score}\left(\vec{u}_m,\vec{c}_{n,t}\right)\right)}{\sum_{m'=1}^{M}exp\left(f_{score}\left(\vec{u}_{m'},\vec{c}_{n,t}\right)\right)}
\label{eq:attention}
\end{equation}
\vspace{-0.5cm}

Where $f_{score}$ is a scoring function used to compute the alignment. This introduces three key considerations for the attention mechanism design:

\vspace{-0.1cm}
\textbf{1. Computing the context}: The context should be a representation of ${H}_{n,1:t}$ that is most appropriate for computing the alignment in Equation \ref{eq:attention}. \cite{SMS-DKL} used a Recurrent Neural Network (RNN) to learn a per-instance representation of a sequence up to some time $t$. Taking the hidden state of an RNN for the context vector enables us to capture correlations: with an RNN, the hidden state is likely to be correlated across time steps, i.e. $\vec{c}_{n,t}\sim \vec{c}_{n,t+h}$  for sufficiently small $h$. Therefore, a model with high weight at $t$ (due to a strong alignment with the context) will most likely have high weight at $t+h$. This is desirable considering that we generally expect a model’s performance to be correlated across time steps (\cite{SMS-DKL}); this is particularly the case with clinical data, where changes in risk factors tend to occur gradually (\cite{HyperLSTM}). We therefore choose to model $f_c$ with a simple RNN.

\vspace{-0.1cm}
\textbf{2. Computing expert encodings}:
For computing per-expert representations $\{\vec{u}_m\}_{m=1}^M$, we largely draw inspiration from \cite{Granger_Causal}: for each expert $m$, we learn a $v$-dimensional vector $\vec{u}_{m}\in\mathbb{R}^v$ representing the context $\vec{c}_{t}$ for which model $m$’s prediction is most useful. These vectors are learned jointly with the base models. We further justify our interpretation of the expert encodings when discussing the MAES training procedure (section \ref{section:MAES_training}).

\vspace{-0.1cm}
\textbf{3. Computing alignments}: Unlike common attention models in NMT, here the set of base learners is not a sequence. Hence the scoring function $f_{score}$ should be purely content-based and invariant to re-ordering for the alignment to remain unchanged under permutations of the base models (Table \ref{table:attention_types}).

\begin{figure*}[!htbp]
    \centering
  {%
    \subfloat[MAES.]{%
      \includegraphics[width=0.65\linewidth]{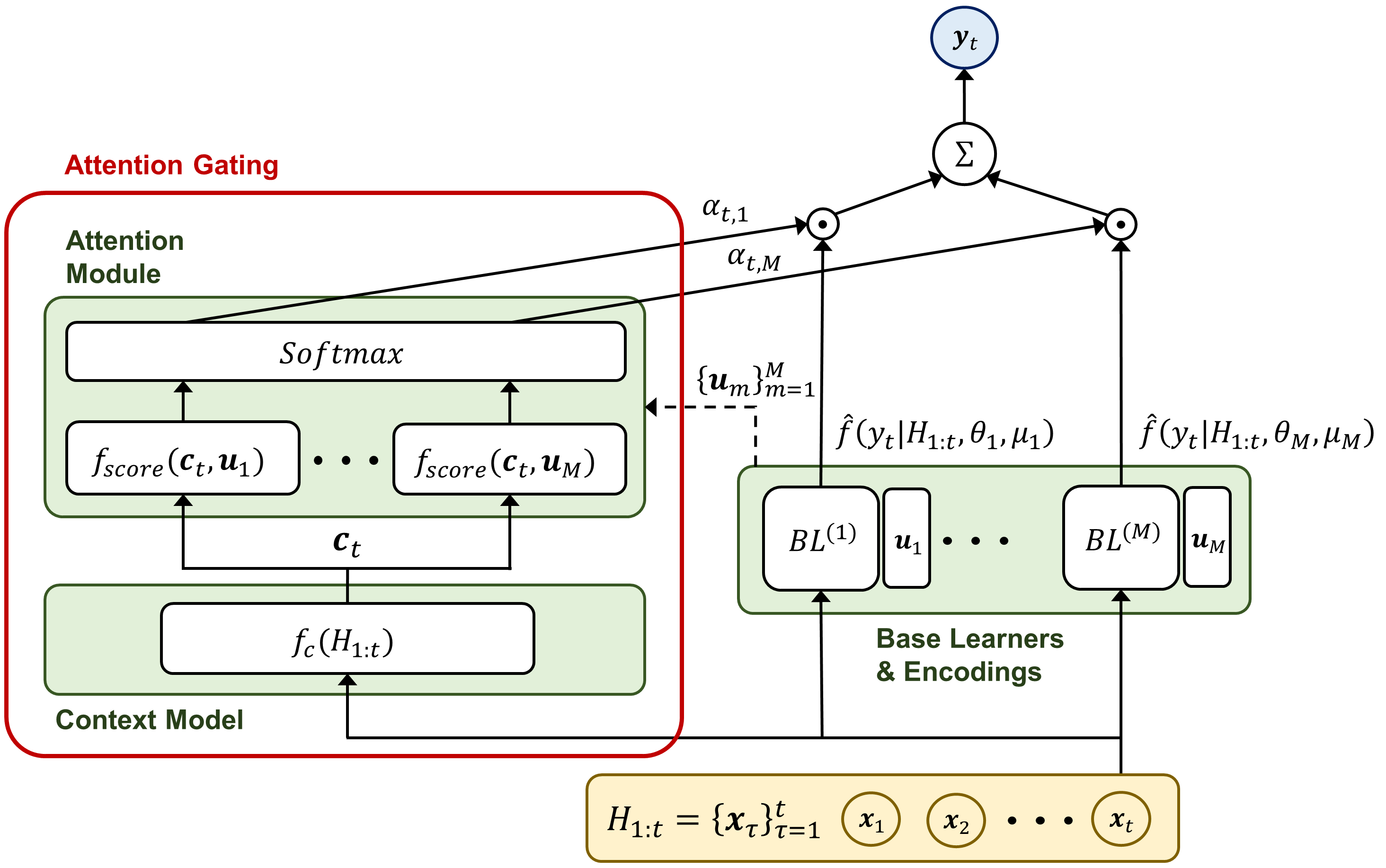}}%
    \qquad
    \subfloat[Stacking Ensemble.]{
      \includegraphics[width=0.28\linewidth]{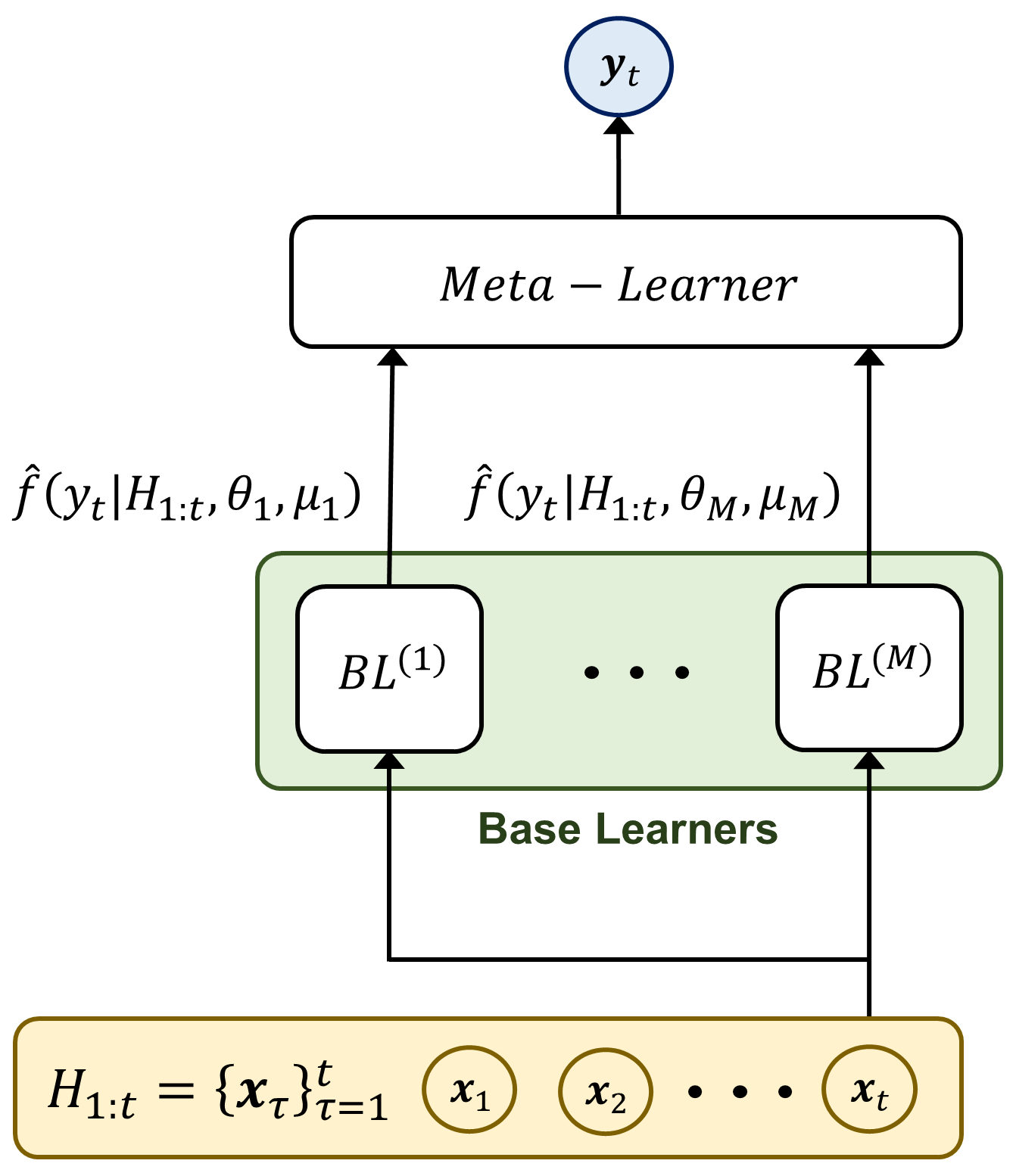}}
  }
    \caption{The full MAES architecture (a), compared to a \textit{classic} stacking ensemble (b). In MAES, each expert issues a prediction given the sequence features up to the prediction step $t$, ${H}_{1:t}=\{\vec{x}_\tau\}_{\tau=1}^T$. In parallel, the context model learns a per-instance representation $\vec{c}_t$ of ${H}_{1:t}$. The attention module computes the alignment between the context and each per-expert encoding $\vec{u}_m$, which represent each expert's ability to model different temporal contexts. The alignment scores are normalized through a softmax function and used to weight the expert predictions. All components are connected and end-to-end trainable. In the stacking ensemble, training occurs in two steps: the models are first trained independently to predict the full target. The meta-learner then uses the base learner predictions on validation data to learn instance-independent aggregation weights (\cite{stackingLSTM}).}
    \label{fig:MAES}
    \vspace{-0.3cm}
\end{figure*}

\vspace{-0.1cm}
The full MAES architecture is shown in Figure \ref{fig:MAES}.

\setlength{\belowcaptionskip}{-2pt}
\begin{table}
\centering
\vspace{0.2cm}
\caption{Popular content-based and invariant attention mechanisms. $\vec{v}$ and $\vec{W}$ are trainable vectors and matrices respectively. In MAES, the context is the query and the expert encodings are the keys. Additive: \cite{NMT_attention_Bahdanau}; Concatenation, Dot, General: \cite{attention_Luong}.}
\begin{tabular}{cc}
\toprule
\multicolumn{1}{c}{\textbf{Attention Type}} & \multicolumn{1}{c}{$f_{score}(\vec{u}_m,\vec{c}_{n,t})$}\\ \midrule
Additive & $\vec{v}^\top\text{tanh}\left(\vec{W}_1\vec{c}_{n,t}+\vec{W}_2\vec{u}_m\right)$ \\
Concatenation & $ \vec{v}^\top\text{tanh}\left(\vec{W}\left[\vec{c}_{n,t};\vec{u}_m\right]\right)$ \\
Dot & $\vec{c}_{n,t}^\top \vec{u}_m$ \\
General & $\vec{c}_{n,t}^\top\vec{W} \vec{u}_m$\\
\bottomrule
\end{tabular}
\label{table:attention_types}
\vspace{0.4cm}
\end{table}

\vspace{-0.3cm}
\subsection{Selecting the Experts}
\vspace{-0.2cm}
MAES provides the flexibility to choose virtually any predictive model for the experts. The per-expert encodings are agnostic to the expert architecture and only the final outputs are needed from the experts when computing the overall prediction. In the context of TS prediction, model heterogeneity can be useful to deal with the varying dynamics (\cite{ADE}). As such, the experts
could be RNNs with different hidden dimensions, or completely different architectures such as RNNs and 1D Convolutional Neural Networks (1D-CNNs).

\subsection{MAES Training}
\label{section:MAES_training}
\vspace{-0.2cm}
All components of MAES are connected and end-to-end trainable. The training procedure must be designed in accordance with our specialization aim, similar to ME. The loss function for MAES is therefore based on the ME loss proposed by \cite{ME_original}, which maximizes the likelihood at the experts' level:

\thinmuskip=0.4mu
\thickmuskip=0.4mu
\scriptspace=0pt

\vspace{-0.5cm}
\begin{equation}
    \mathcal{L} = -\sum_{n=1}^N\sum_{t=1}^{T}\log\sum_{m=1}^M\alpha_{t,m}^n\prod_{k=1}^K{\hat{f}^k({H}_{n,1:t};\theta_m,\mu_m)}^{\vec{y}_{n,t}^k}
    \label{eq:MAES_loss}
\end{equation}
\vspace{-0.4cm}

\thinmuskip=3mu
\thickmuskip=5mu
\scriptspace=0.5pt

Together with the instance-dependent gating architecture, this loss function encourages specialization: the gate is trained to assign a high weight to well-performing experts on a given context (experts with high likelihood), and the strength of each expert's parameter update is proportional to their associated weight. In other words, experts which perform well on a given context are assigned a higher weight and receive stronger updates for the given instance, thereby specializing on the context's conditional distribution.

Having defined our loss function, we provide further intuition into the expert encodings $\{\vec{u}_m\}_{m=1}^M$. For a well-performing expert $m$ to be assigned a high weight on a given context, the alignment between the context and the expert encoding $\vec{u}_m$ must be maximized. The expert encoding is therefore trained to best align with the contexts on which its associated expert performs well. In essence, $\vec{u}_m$ represents the context $\vec{c}_t$ for which model $m$'s contribution should be maximal.

\vspace{-0.2cm}
\section{Experiments}
\vspace{-0.2cm}
We decided to evaluate the performance of MAES on synthetic datasets, as they enable us to control the amount of TCS.

\vspace{-0.3cm}
\subsection{Methods}
\vspace{-0.3cm}
\subsubsection{Datasets}
\vspace{-0.2cm}
\label{section:datasets}
The synthetic datasets were generated by adapting the procedure outlined in \cite{HyperLSTM} for equal-length input and output sequences with binary targets. The datasets consist of $N$ sequences of $T$ input features $\{\vec{x}_t\}_{t=1}^T$, $\vec{x}_t\in \mathbb{R}^d$, and their corresponding targets $\{{y}_t\}_{t=1}^T$, $y_t\in \{0,1\}$. The target at each time step is a weighted combination of the features in a window of $l$ previous time steps (with a sigmoid function applied at the end, to generate binary labels), where two sets of weights $\vec{w}^{(l)}\in\mathbb{R}^l$ and $\vec{w}^{(d)}\in\mathbb{R}^d$ dictate the conditional relationship across time steps and across feature dimensions respectively. TCS is simulated by altering the weights $\vec{w}^{(l)}$ and $\vec{w}^{(d)}$ across time steps:

\vspace{-0.3cm}
\begin{equation}
    y_t = \sigma\left(\vec{w}^{(l)\top}_{t} [\vec{x}_{t-l},\ldots,\vec{x}_{t-1}]^\top\vec{w}^{(d)}_{t}\right)
\end{equation}
\vspace{-0.4cm}

Where $\vec{w}_t^{(\cdot)} = \vec{w}_{t-1}^{(\cdot)} + \Delta_t$ for $\Delta_t$ sampled from $\Delta_t\sim \text{Uniform}(-\delta,\delta)$. The parameter $\delta$ modulates the amount of variability in the parameters of the data generation process; $\delta$ therefore controls the amount of temporal shift.

We generated datasets for $\delta\in \{0.0,\,0.01, 0.025,\,0.05,\,0.075,\,0.1,\,0.2,\,0.3,0.4\}$, each containing $N_{train}=5,000$ and $N_{test}=1,000$ sequences for training and testing respectively. We set aside $20\%$ of each training dataset for validation and best model saving. For each dataset, we sampled sparse inputs and set the ratio of positive labels was set to $r=0.25$ to simulate label imbalance often arising in healthcare datasets. In all experiments, we used $d=3$, $l=10$ and $T=48$.

\vspace{-0.3cm}
\subsubsection{Baseline Models}
\vspace{-0.2cm}
As stated previously, recurrent models struggle to adapt to TCS due to their complete parameter sharing over time. To verify this and investigate the potential benefits of ensemble learning methods for TCS adaptation with `classic' recurrent models, we evaluated the performance of heterogeneous LSTMs with different memory capacities, both when trained and evaluated individually and when used as part of an ensemble.

We randomly sampled $M=20$ LSTM hidden dimensions $h_{dim}$ from the range $h_{dim}\in[100,1100]$. All $20$ LSTMs have a single recurrent layer with tanh activation and a time-distributed feed-forward layer with a sigmoid activation at the output, to produce class probabilities at each time step. The following baseline models were constructed from these $M$ base learners:

\begin{enumerate}[wide, labelwidth=!, labelindent=0pt]
\vspace{-0.2cm}
    \item \textbf{Individual Models}. Training and evaluating the LSTMs independently enabled us to demonstrate their limitations with TCS.
    \vspace{-0.1cm}
    \item \textbf{Post-hoc step-wise selection}. The second baseline is similar to SMS-DKL (\cite{SMS-DKL} -- Section \ref{section:related_work}): at each time step, we select the predictions of the LSTM with the lowest validation loss, to assess the benefits of step-wise model selection and compare it to step-wise EL (baseline 3.c.). If we denote $\mu^{*}(t)\in \{\mu_m\}_{m=1}^{M}$ the model hyperparameters with minimum validation loss at time $t$, this corresponds to finding the set of $T$ architectures $\{\mu^{*}_t\}_{t=1}^T$ that minimize the step-wise validation loss:
    
    \thinmuskip=2mu
\thickmuskip=2mu
\scriptspace=0.5pt

\vspace{-0.5cm}
    \begin{equation}
        \{\mu^{*}_t\}_{t=1}^T= \argminB_{\{\mu_t\}_{t=1}^T} \sum_{n=1}^N\sum_{t=1}^T \mathcal{L}\left(y_{n,t},\hat{f}\left(H_{n,t}^{val};\theta,\mu_t\right)\right)
    \end{equation}
\vspace{-0.2cm}

\thinmuskip=3mu
\thickmuskip=5mu
\scriptspace=0.5pt
    
    Where $\mathcal{L}((y_{n,t},\hat{f}(H_{n,t}^{val};\theta,\mu_t))$ is the validation loss at prediction step $t$ for the model with hyperparameters $\mu_t\in \{\mu_m\}_{m=1}^M$. The difference with \cite{SMS-DKL} is that in their work, a BO procedure was developed to identify the optimal model for each time step across a wide range of possible hyperparameters, while here we select the hyperparameters among the $M$ individual LSTM architectures.
    
    \vspace{-0.1cm}
    \item \textbf{Ensemble Learning}. Three popular EL techniques were evaluated: 
    \begin{enumerate}[wide, labelwidth=!, labelindent=0pt]
    \vspace{-0.1cm}
        \item \textbf{Average ensemble}. This is the most primitive type of ensemble method, where the predictions of different models are averaged, but it has often been shown to achieve surprisingly good performance (\cite{ADE}):
        
        \vspace{-0.45cm}
    \begin{equation}
        p(\vec{y}_{n,t}|H_{n,1:t})=\frac{1}{M}\sum_{m=1}^M \hat{f}\left({H}_{n,1:t};\theta_m,\mu_m\right)
    \end{equation}
        \vspace{-0.35cm}
        
        \item \textbf{Global stacking ensemble}. The model predictions are weighted according to $M$ \textit{global} weights $\{w_{m}\}_{m=1}^M$ learned by a linear meta-learner on validation data (\cite{Stacking_original}):
        
    \vspace{-0.55cm}
    \begin{equation}
        p(\vec{y}_{n,t}|H_{n,1:t})=\sum_{m=1}^Mw_m \hat{f}\left({H}_{n,1:t};\theta_m,\mu_m\right)
    \end{equation}
    \vspace{-0.35cm}

    \item \textbf{Step-wise stacking ensemble}. To incorporate some time-adaptation in the ensemble predictions, the linear meta-model can learn $T$ independent weights $\{w_{m,t}\}_{t=1}^T$ for each model $m$ instead:
    
    \vspace{-0.45cm}
    \begin{equation}
        p(\vec{y}_{n,t}|H_{n,1:t})=\sum_{m=1}^Mw_{m,t} \hat{f}\left({H}_{n,1:t};\theta_m,\mu_m\right)
    \end{equation}
    \vspace{-0.4cm}
    
    This corresponds to the method developed in \cite{stackingLSTM} (Section \ref{section:related_work}); while the authors also tested more complex meta-models such as random forests, the linear meta-learner generally performed better in their study.
    \end{enumerate}
\end{enumerate}

\vspace{-0.5cm}
\subsubsection{Model Training and Configurations}
\vspace{-0.2cm}
All models were trained for up to $15$ epochs using the Adam optimizer (\cite{Adam}). The default values of $\beta_1=0.9$ and $\beta_2=0.999$ were used for Adam, and the batch size and learning rate were set to $100$ and $0.001$ respectively, as in \cite{HyperLSTM}. The baselines were all trained with a standard Binary Cross Entropy (BCE) loss. The stacking ensembles' meta-models were trained on the validation data, for $1,000$ gradient descent steps.

For MAES, a single-layer RNN context model was used for computational simplicity. An ablation study was performed on the $\delta=0.2$ validation data to identify the optimal training procedure, attention model architecture and number of base learners (cf. supplementary material). This exploratory analysis highlighted the superior performance of our bespoke loss function from Equation \ref{eq:MAES_loss}. The best-performing MAES architecture consisted of $M=5$ experts (randomly sampled among the $20$ LSTMs) and an Additive attention mechanism for the gate. The optimal architecture and training procedure were maintained when evaluating MAES on all datasets.

\vspace{-0.3cm}
\subsubsection{Model Evaluation}
\vspace{-0.2cm}
The Area under the Precision-Recall curve (APR) was used to evaluate the models, since it is a threshold-invariant metric that accounts for the distribution skew in the datasets. The APR was computed at every step and averaged across the sequence length to yield a single performance measure for each model, as in \cite{SMS-DKL}.

\vspace{-0.3cm}
\subsection{Results}
\vspace{-0.2cm}

\begin{figure}[h]
\begin{minipage}{\linewidth}
  \includegraphics[width=0.97\linewidth]{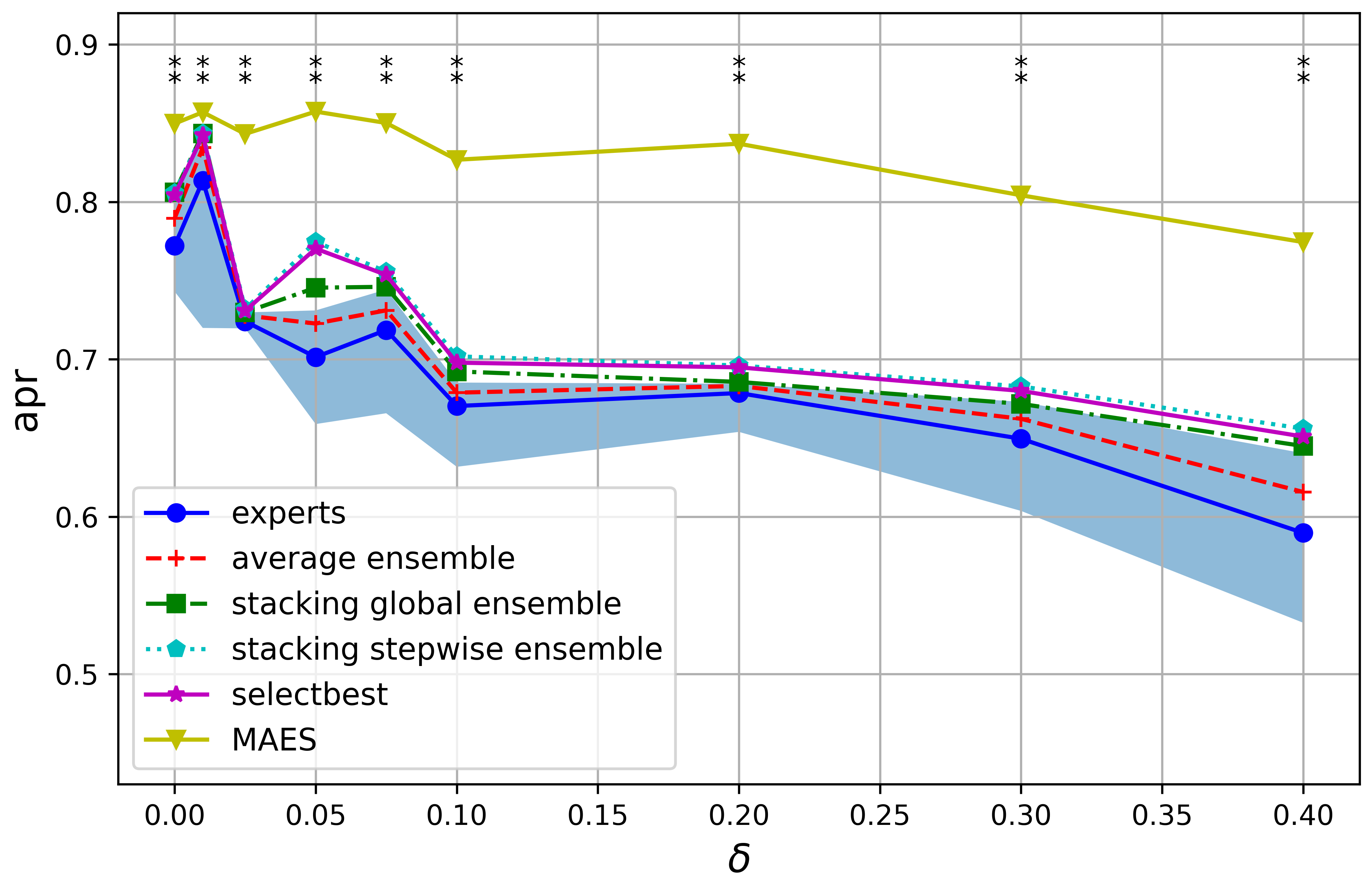}
\caption{Model APR for different amounts of temporal conditional shift.\footnotemark \,The shaded area encompasses all the individual LSTMs (`experts'). MAES' performance is significantly more robust to increasing amounts of temporal shift than all baselines.}
\label{fig:apr_all}
\end{minipage}
\end{figure}

The APR of all models for increasing amounts of TCS is shown in Figure \ref{fig:apr_all}.

\setlength{\belowcaptionskip}{0pt}

\setlength{\belowcaptionskip}{0pt}
\begin{figure}[!htbp]
\begin{minipage}{\linewidth}
\centering
\vspace{0.05cm}
\includegraphics[width=0.93\linewidth]{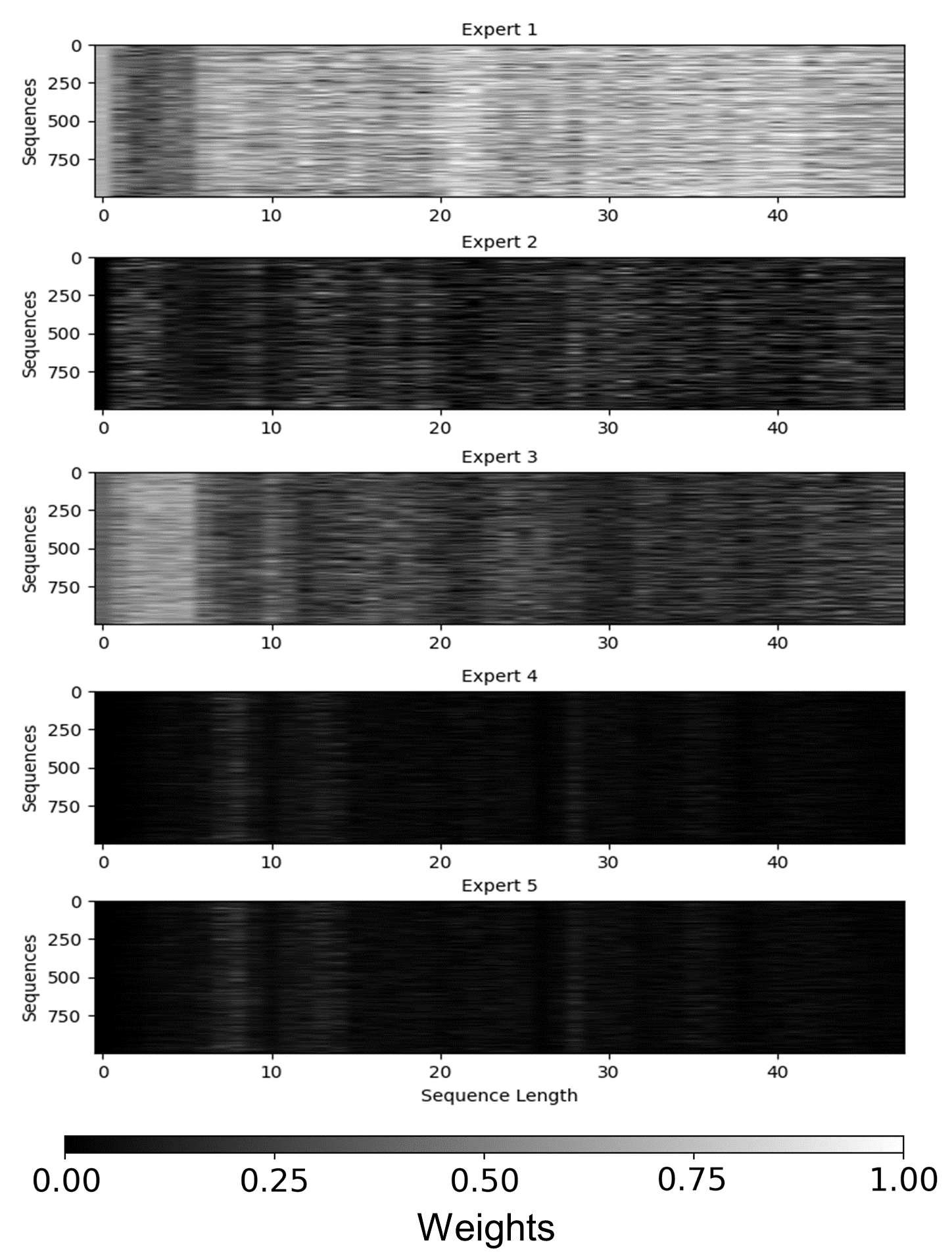}
\caption{Attention weights for all MAES experts, for each sequence in the test set with $\delta=0.2$. The experts are LSTMs with randomly sampled hidden dimensions $h_{dim}=[700, 400, 420, 360, 260]$ (in order). Through MAES' attention gating mechanism, the expert weights depend on both the specific sequence features and the prediction step.}
\label{fig:attention_weights_expert_all}

\vspace{0.19cm}
\footnotetext{$^1$Significance levels, reported at p=0.2 (*) and p=0.05 (**), were computed for MAES and the best-performing baseline using a Monte Carlo permutation test.}
\end{minipage}
\end{figure}

\setlength{\belowcaptionskip}{-10pt}
\setlength\intextsep{0pt}
\begin{figure*}[t]
\centering
  {%
    \subfloat[Step-wise stacking ensemble predictions. The base learners' predictions are highly correlated. The red shading encompasses \textit{all} base learner predictions.\label{fig:baseline_predictions}]{%
      \includegraphics[width=0.8\linewidth]{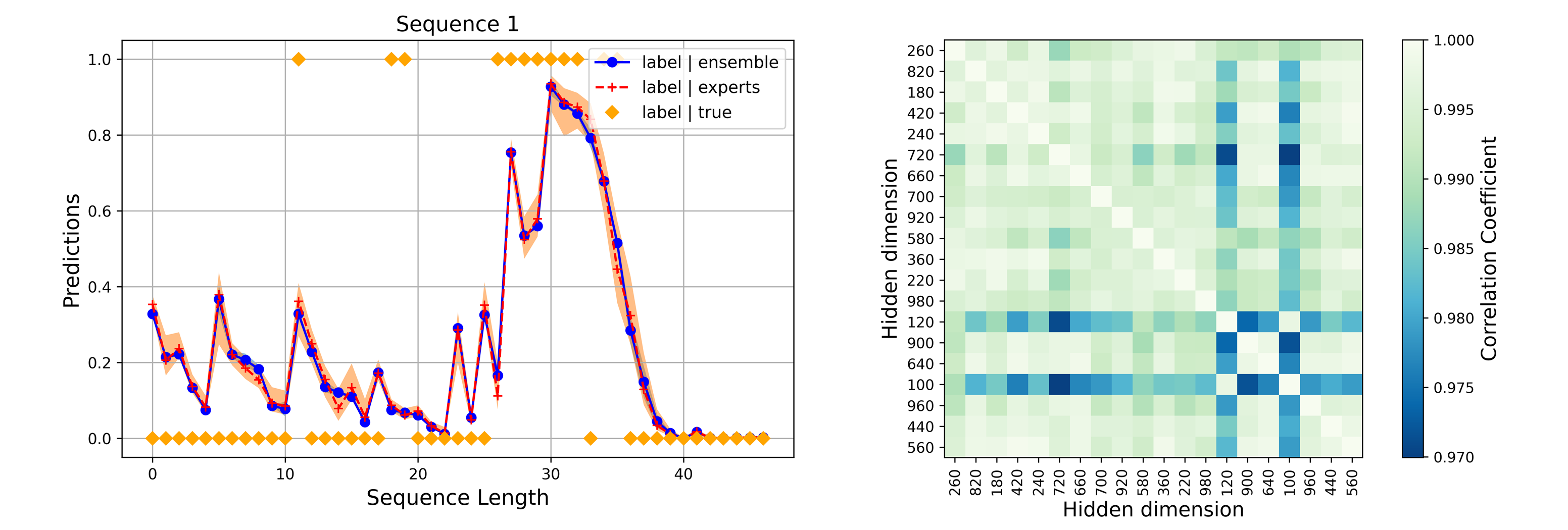}}%
    \qquad
    \subfloat[MAES predictions. Specialization in MAES results in predictions with low correlation. The uncertainties in the ensemble predictions (blue shading) were computed from the weighted standard deviation in base learner predictions.\label{fig:MAES_predictions}]{
      \includegraphics[width=0.8\linewidth]{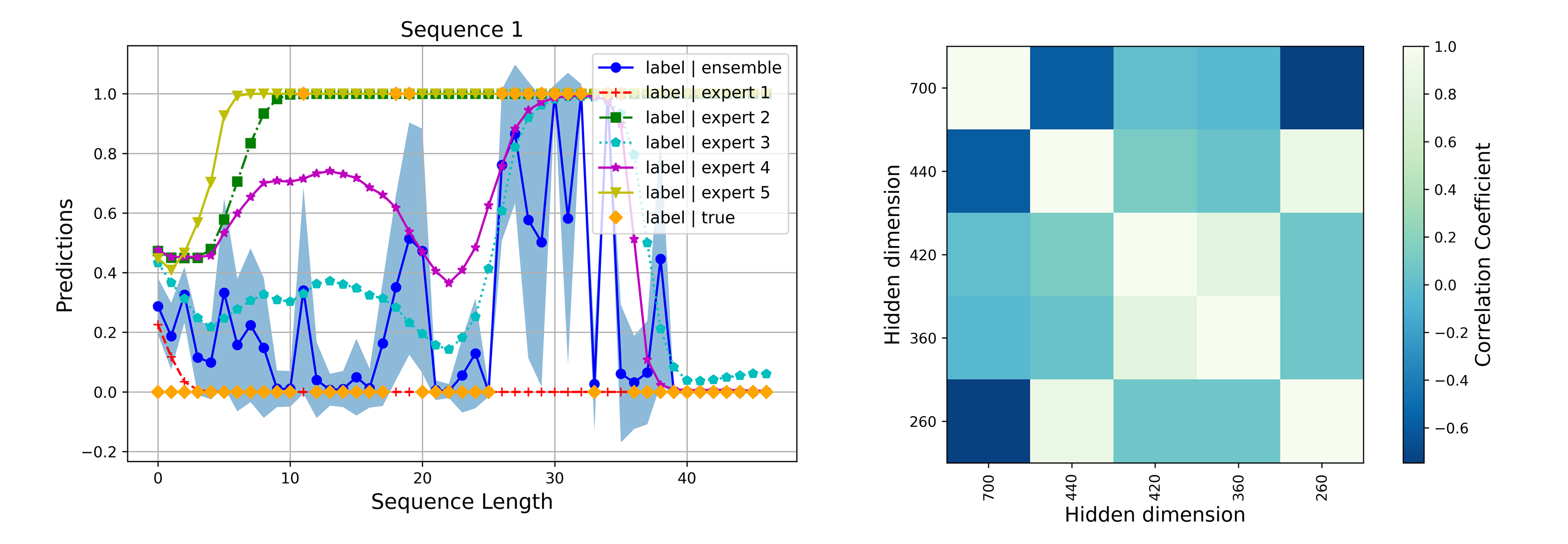}}
  }
    {\caption{Step-wise stacking and MAES predictions for a sample test sequence (left), along with the base learners' correlation plots (right, with their associated hidden dimension on the axes). The correlation heatmaps were obtained by computing the pairwise Pearson correlation coefficient between the base learners' predictions. Note that in MAES, only $5$ experts were maintained from the ablation study. \label{fig:prediction_plots}}}
\end{figure*}

As expected, step-wise stacking and post-hoc step-wise selection both helped improve performance relative to their global counterparts (global stacking and single LSTMs respectively) for increasing TCS: the contributions of the base models vary along the sequence, with each base learner potentially modeling a different conditional distribution due to their heterogeneity. The baseline models also benefited from ensembling overall: for all $\delta$, global and step-wise stacking outperformed individual LSTMs and post-hoc step-wise selection respectively. The average ensemble achieved lower APR values than the best base models, as the under-performing base learners contribute equally to the ensemble prediction.

These differences in performance are still relatively limited. MAES, however, is significantly more robust to TCS than all baselines as its APR remains fairly stable with increasing TCS. The performance improvements can be interpreted from the sample predictions of Figure \ref{fig:prediction_plots}. Despite their different hidden dimensions, all base learners issue highly correlated predictions when trained independently. As such, the potential improvements of baseline ensemble methods are minimal. With MAES, however, all experts issue different predictions over the same sequence as a consequence of their specialization (Figure \ref{fig:MAES_predictions}) and their relative contributions vary across sequences to enable MAES to adapt to the shifting dynamics (Figure \ref{fig:attention_weights_expert_all}). In fact, all experts have regimes with non-negligible weights that vary according to the experts' predictive ability on the temporal context (Figure \ref{fig:attention_weights_expert_all}).

\vspace{-0.1cm}
Another source of improvement for MAES is the instance-dependent weighting of base learner predictions (Figure \ref{fig:attention_weights_expert_all}): the attention weights depend on both the time step and the instance features (mainly on time due to the predominance of TCS). Figure \ref{fig:attention_weights_expert_all} also demonstrates that in MAES, experts with larger hidden dimensions tend to contribute more to the ensemble prediction, most certainly due to their ability to capture more complex time-varying relationships given their larger memory capacity.

\vspace{-0.1cm}
Future work will further investigate the model's capabilities on \textit{real} data and develop an optimal selection procedure for the base learners, which were randomly sampled in this study. Various design choices will also be investigated, such as the use of entirely different base model architectures (e.g. LSTMs and 1D-CNNs).

\vspace{-0.35cm}
\section{Conclusion}
\vspace{-0.35cm}
The omnipresence of temporal conditional shift in medical sequential datasets prompted for the development of adaptive sequence models. In this work, we highlighted the limitations of popular time-series models in adapting to distribution shifts. We designed and evaluated a novel ensemble learning method that is significantly more robust to rapidly varying dynamics, therefore showing tremendous promise in enabling reliable and personalized predictions of patient outcomes.

\clearpage
\subsubsection*{Acknowledgements}
This work was supported by The Alan Turing Institute (ATI) under the EPSRC grant EP/N510129/1.
\bibliography{MAES}

\end{document}

\onecolumn
\aistatstitle{Model-Attentive Ensemble Learning for Sequence Modeling: \\
Supplementary Materials}
\begin{samepage}
\section{LSTM Parameter Sharing and Conditional Shift}
The LSTM is an extension to the Recurrent Neural Network designed to better capture long-term dependencies through a gated memory mechanism (\cite{LSTM}). The latter consists of a hidden state $\vec{h}_t$ and a cell state $\vec{c}_t$, which are updated according to the following equations:
\vspace{-0.1cm}
\begin{align}
    \vec{i}_t&=\sigma(\vec{W}_{ix}\vec{x}_t+\vec{W}_{ih}\vec{h}_{t-1}+\vec{b}_i)\\
    \vec{f}_t&=\sigma(\vec{W}_{fx}\vec{x}_t+\vec{W}_{fh}\vec{h}_{t-1}+\vec{b}_f)\\
    \vec{o}_t&=\sigma(\vec{W}_{ox}\vec{x}_t+\vec{W}_{oh}\vec{h}_{t-1}+\vec{b}_o)\\
    \Tilde{\vec{C}}_t&=\text{tanh}(\vec{W}_{\Tilde{c}x}\vec{x}_t+\vec{W}_{\Tilde{c}h}\vec{h}_{t-1}+\vec{b}_{\Tilde{c}})\\
    \vec{C}_t&=\sigma(f_t\odot \vec{C}_{t-1}+i_t\odot\Tilde{C}_t)\\
    \vec{h}_t&=\text{tanh}(\vec{C}_t)\odot\vec{o}_t
\end{align}
\vspace{-0.75cm}

Where $\odot$ denotes an element-wise multiplication. A prediction can be issued by adding an appropriate output layer (e.g. a fully-connected linear layer for regression) on top of the hidden state $\vec{h}_t$.

The issue in Equations 1--6 is that the parameters $\vec{W}_{(\cdot)}$ and $\vec{b}_{(\cdot)}$ are {\textit{fixed}} across time steps. As such, the varying dynamics of datasets subject to temporal conditional shift must be implicitly incorporated in the finite-length hidden and cell states (\cite{HyperLSTM}). Since LSTMs are already hard to train, such temporal variations are generally not captured by these models and special adaptation methods must be considered to explicitly capture temporal shift. We discuss these adaptations in our paper (Section 3).

\section{Bayesian Perspective of Ensemble Learning}\label{apd:ensemble_learning}
The design of MAES was derived from a Bayesian approach to supervised learning. Let us assume that we have a dataset with $N$ sequences $\mathcal{D}=\{\vec{s}_n,\vec{x}_{n,1:T_n}\}_{n=1}^N$ along with the corresponding label sequences $\{\vec{y}_{n,1:T_n}\}_{n=1}^N$, and the task is to estimate a predictive distribution $p(\vec{y}_{n,t}|H_{n,t})$ for a test sequence $H_{n,t}=(\vec{s}_n,\vec{x}_{n,1:t})$.

A standard model selection approach generally involves:

\vspace{-0.4cm}
\begin{enumerate}
    \item Choosing the model hyperparameters $\mu_{m}\in\mathcal{M}$, where $\mathcal{M}$ is the set of all possible hyperparameters. Note that here we use the term \textit{hyperparameters} in a broad sense which includes both the architecture and the specific hyperparameters for the architecture.
    \vspace{-0.2cm}
    \item Finding the best set of model parameters $\theta_{m}^*\in\Theta_{m}$ on the training data $\mathcal{D}$ (e.g. through maximum-likelihood estimation), where $\Theta_{m}$ is the parameter space for the chosen model architecture.
    \vspace{-0.2cm}
    \item Using the trained model to issue a prediction $p(\vec{y}_{n,t}|H_{n,t})=\hat{f}(H_{n,1:t};\theta_{m}^*,\mu_{m})$.
\end{enumerate}
\vspace{-0.3cm}

This process can easily lead to hypotheses with high bias and variance. The selected model hyperparameters might not have the appropriate hypothesis space for the problem, and might under-perform on some subsets of the data. The model parameters can also overfit the training data, thus leading to high variance. Most importantly, parameter sharing in most single-model techniques may lead to high temporal biases in the presence of temporal conditional shift. Instead, we could consider the space of all possible models $\mathcal{M}$ and their parameters $\theta$. In this case, the predictive distribution can be expressed as:

\vspace{-0.3cm}
\begin{equation}
    p(\vec{y}_{n,t}|H_{n,1:t})=\sum_{\mu\in\mathcal{M}}\int p(\theta,\mu|H_{n,1:t},\mathcal{D})p(\vec{y}_{n,t}|{H}_{n,1:t},\theta,\mu)d\theta
\label{eq:bayesian_EL}
\end{equation}
\vspace{-0.4cm}

\end{samepage}

Ensemble learning methods generally use a Monte Carlo approximation to the above integral:

\vspace{-0.3cm}
\begin{equation}
p(\vec{y}_{n,t}|{H}_{n,1:t})=\sum_{m=1}^M w_m p(\vec{y}_{n,t}|{H}_{n,1:t},\theta_m,\mu_m)
\label{eq:ensemble_learning_TS_simple}
\end{equation}
\vspace{-0.3cm}

Where the sampled models are assigned weights $\{w_m\}_{m=1}^M$ according to the ``trust'' in their predictions, generally obtained from the models' validation performance or by training a meta-model on their predictions (stacking). These weights essentially emulate $p(\theta,\mu|\mathcal{D})$ as they are often assumed to be independent of the test instance $H_{n,1:t}$ (when using validation performance or stacking for example).

In MAES, we want the model distribution to depend on time and on the speciﬁc input features: $\{w_m\}_{m=1}^M\sim p(\theta,\mu|H_{n,1:t},\mathcal{D})$. We arrive at the following predictive distribution for MAES:

\vspace{-0.3cm}
\begin{equation}
p(\vec{y}_{n,t}|{H}_{n,1:t})= \sum_{m=1}^M w_m(H_{n,1:t})p(\vec{y}_{n,t}|{H}_{n,1:t},\theta_m,\mu_m)
\label{eq:MAES_eqn}
\end{equation}
\vspace{-0.4cm}

Where we made explicit the weights' dependence on the test instance. With instance-dependent weights, Equation \ref{eq:MAES_eqn} most closely approximates Equation \ref{eq:bayesian_EL} for a finite number of models $M$.

\section{MAES Ablation Study}
\label{apd:MAES_exploratory_analysis}
There are many design considerations and testing conditions to investigate in MAES, such as the choice of attention mechanism or the effect of the number of experts on MAES performance. Prior to comparing MAES to different baselines for various amounts of temporal conditional shift (Section 5.2), we evaluated MAES on validation data for a dataset with a fixed intermediate amount of conditional shift $\delta=0.2$, under different conditions and for various architecture choices.

Specifically, we focused on identifying (1) the optimal training procedure, (2) the optimal scoring function in the attention mechanism and (3) the optimal number of experts. In all experiments, a single layer was used for all neural-network-based scoring functions along with a single-layer RNN context model for computational simplicity. For each MAES configuration investigated, a random search was performed over the hidden dimension of the context model, the hidden dimension of the attention mechanism (where applicable) and the per-expert encodings' dimension. We uniformly sampled $20$ different values from the range $\{10, 30, \ldots, 1100\}$ for each of the $3$ aforementioned hyperparameters.

\subsection{MAES Training Procedure}

We initially wanted to review the MAES training procedure developed in Section 4.3 and compare the loss function $\mathcal{L}$ of Equation 12 (main paper) to a more \textit{classic} BCE loss. We first trained MAES with a cross-entropy loss function $\mathcal{L}_{BCE}$. We then trained the model $\mathcal{L}$, where we use the individual expert predictions directly.

One issue that arises frequently in mixtures of experts is that the gate tends to assign high weights to the select few experts that perform well in the initial stages of training (\cite{ME_loss_Eigen}). As these experts perform well, the gate assigns them a higher weight and their parameters are updated faster, thus causing them to perform better and to be selected more frequently on subsequent examples. The presence of such local optima is further reinforced by the loss function $\mathcal{L}$ being non-convex. While sampling different architectures (as done in our random sampling procedure) can help circumvent these issues, we investigated two additional methods:
\vspace{-0.2cm}
\begin{enumerate}[wide, labelwidth=!, labelindent=0pt]
    \item \textbf{Importance loss}. To tackle the self-reinforcing problem outlined above, \cite{Outrageously_large_MoE} defined an additional \textit{importance} loss $\mathcal{L}_{imp}$, which was added to the overall loss with a tunable scaling parameter $w_{imp}$:
    \vspace{-0.2cm}
    \begin{equation}
        \mathcal{L}_{tot} = \mathcal{L} + w_{imp}\mathcal{L}_{imp}
    \end{equation}
    \vspace{-0.7cm}

    We drew inspiration from \cite{Outrageously_large_MoE} and defined our importance loss as:
    \vspace{-0.2cm}
    \begin{equation}
        \mathcal{L}_{imp} = -\sum_{m=1}^M\left(\sum_{n=1}^N\sum_{t=1}^T\alpha_{t,m}^n\right)^2
    \end{equation}
    \vspace{-0.4cm}

    This loss encourages the attention weights to be more evenly distributed across the experts. The importance coefficient was tuned through a grid-search on the validation data with $w_{imp}\in\{0,0.1,\ldots,1\}$.
    
    The rationale for including an importance loss factor can also be derived from the ME trade-offs between model fusion and selection (\cite{ME_review}). By specialising models on the input space, ME tends to assign a high weight to few experts for a given instance (\textit{selection}). However, as in most ensemble learning methods, the purpose is also to group contributions from more than one expert e.g. for variance reduction (\textit{fusion}). The importance loss thus enables us to find the optimal balance between model selection and fusion.
    
    \vspace{-0.2cm}
    \item \textbf{Expert pre-training}. Pre-training the experts individually for a certain number of epochs ensures that all expert parameters are initially updated. We therefore trained the experts separately for different numbers of epochs, loaded their parameters in MAES and trained the full model jointly for the remaining number of epochs for a total of $15$ epochs (pre-training + MAES training).
\end{enumerate}
\vspace{-0.1cm}

\begin{table}[!htbp]
\caption{MAES validation performance for different training procedures.}
\centering
\begin{tabular}{lc}
\toprule
\multicolumn{1}{c}{\textbf{Model}} & \textbf{APR} \\ \midrule
BCE & $0.8414\pm0.1553$ \\
$\mathcal{L}$ ($w_{imp}=0.0$) & $\vec{0.8433\pm0.1534}$ \\ \bottomrule
\end{tabular}
\label{table:synthetic_MAES_training_results}
\end{table}

We found that training MAES with our loss function $\mathcal{L}$ resulted in an improved performance relative to the simple BCE loss (Table \ref{table:synthetic_MAES_training_results}). $\mathcal{L}$ was designed to mimic the ME loss, forcing each expert to predict the complete target and thereby limiting the risk of overfitting compared to BCE training while encouraging the experts to specialize.

\vspace{-0.1cm}
\begin{figure}[!htbp]
    \centering
  {%
    \subfloat[Varying the importance factor $w_{imp}$, designed to encourage weights to be more distributed across experts. \label{fig:adaptive_wimp}]{%
      \includegraphics[width=0.47\linewidth]{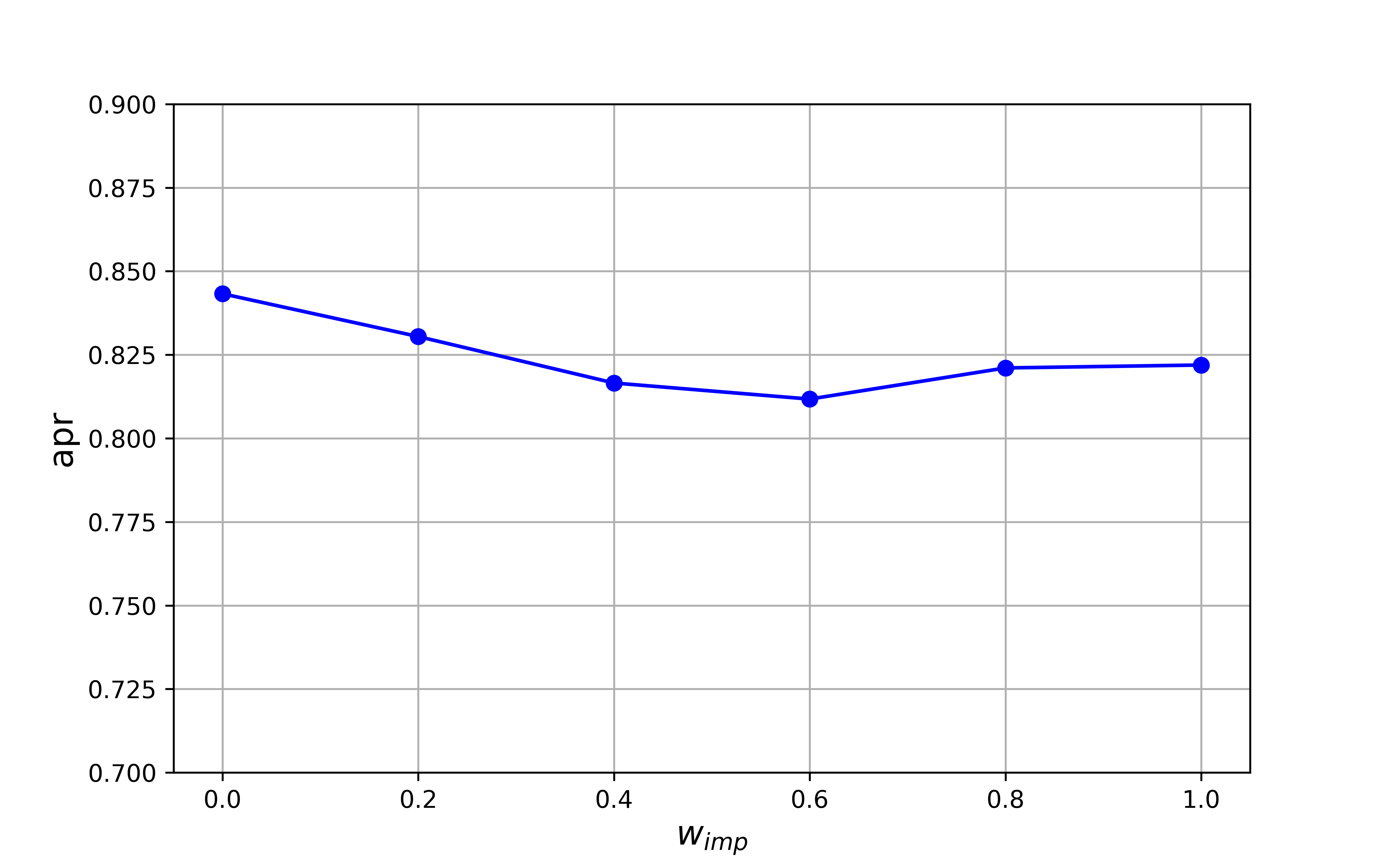}}%
    \qquad
    \subfloat[Pre-training experts to ensure that all of their parameters are initially updated. \label{fig:adaptive_pretrain}]{
      \includegraphics[width=0.47\linewidth]{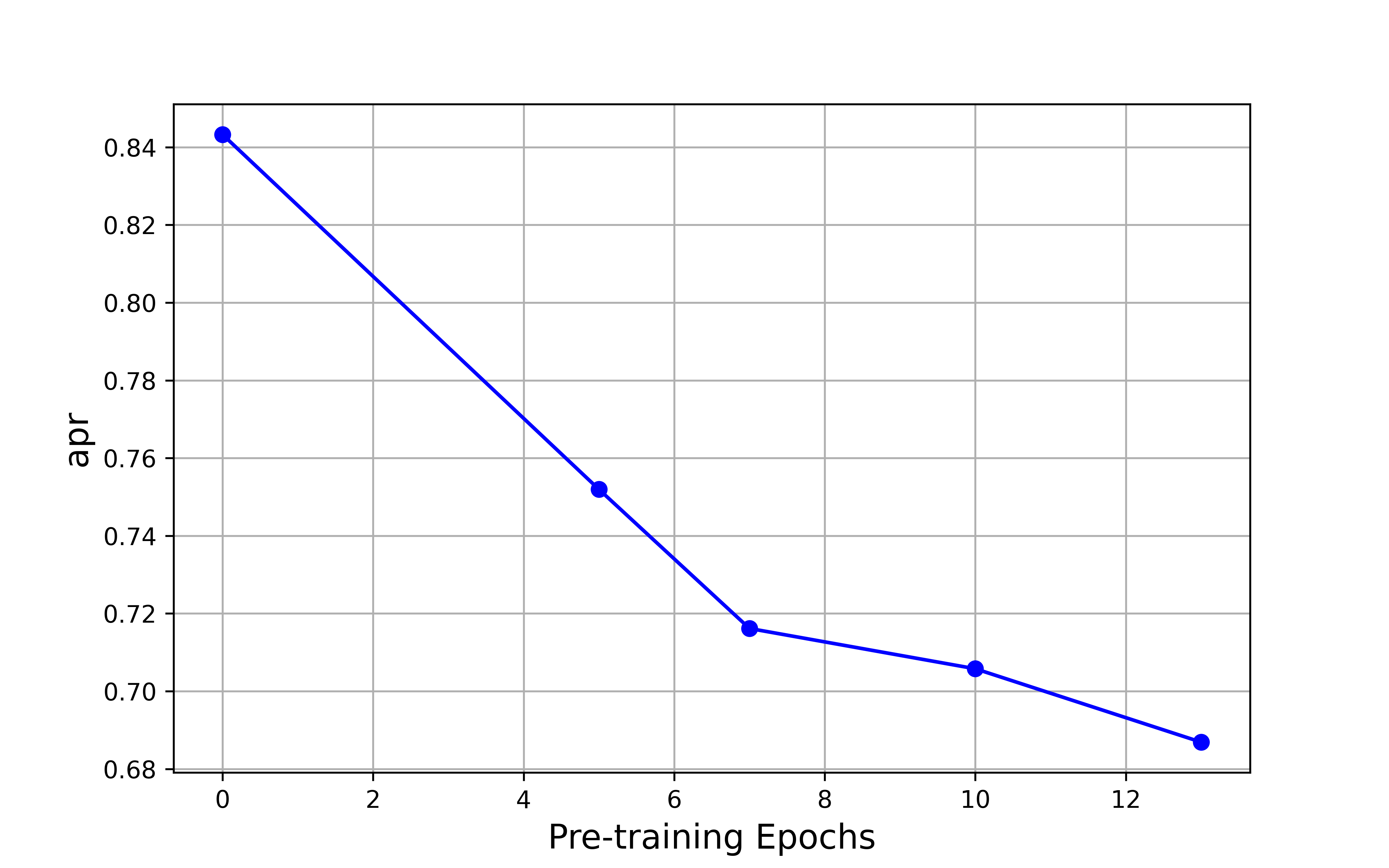}}
  }
    \caption{Validation performance of MAES on the $\delta=0.2$ dataset with different training methods. (a) Increasing the importance loss did not improve performance, as the optimal balance between model selection and fusion is achieved for $w_{imp}=0.0$. (b) While pre-training the experts ensures that all of their weights are initially updated, this is inconsistent with the specialization aim of MAES and results in a decrease in performance.}
    \label{fig:ablation_results}
\end{figure}

Incorporating an importance loss factor did not improve model performance here -- the optimal balance between model selection and fusion is achieved for $w_{imp}=0$ (Figure \ref{fig:adaptive_wimp}). Pre-training the experts also resulted in a degradation in performance (Figure \ref{fig:adaptive_pretrain}). The pre-training procedure is inconsistent with the specialization aim of MAES: experts learn to model the average trend in the data in the initial training phase, and the gate then attempts to specialize them for the remaining epochs.

In the subsequent sections, we used the best-performing training procedure for MAES: a bespoke loss $\mathcal{L}$ with $w_{imp}=0.0$.

\subsection{Attention Mechanism}

We evaluated the use of the content-based scoring functions listed in Table 2 (main paper) for the attention module. The results are shown in Table \ref{table:attention_results}.

Additive and Concatenation attention performed similarly. In both cases, the alignment is computed using a neural network: in the former, a separate set of weights is used for the key (context) and queries (per-expert encodings) which are then added together, while in the latter both the key and query are concatenated at the neural network input. Dot attention is a more simple mechanism, which does not use any trainable parameters. It is therefore more data-efficient but less flexible, and performed slightly less well than the neural-network-based architectures. Note that for Dot attention, we had to ensure that the per-expert encodings and the context vector had the same dimension in the random search over these two hyperparameters. General attention is a flexible extension of Dot attention. Here, however, General attention resulted in a lower APR, which might be due to the relatively limited number of sampled architectures in the random search.

While no single architecture performed significantly better, the slightly superior Additive attention model was maintained in subsequent experiments.

%\vspace{-0.2cm}
\begin{table*}[htbp]
\caption{MAES validation performance for different attention models.}
\centering
\begin{tabular}{lcc}
\toprule
\multicolumn{1}{c}{\textbf{Attention Type}} & \multicolumn{1}{c}{$f_{score}(\vec{u}_m,\vec{c}_{n,t})$} & \multicolumn{1}{c}{\textbf{APR}} \\ \midrule
Additive & $\vec{v}^\top\text{tanh}\left(\vec{W}_1\vec{c}_{n,t}+\vec{W}_2\vec{u}_m\right)$ & $\vec{0.8433\pm0.1534}$ \\
Concatenation & $ \vec{v}^\top\text{tanh}\left(\vec{W}\left[\vec{c}_{n,t};\vec{u}_m\right]\right)$ & $0.8363\pm 0.1520$ \\
Dot & $\vec{c}_{n,t}^\top \vec{u}_m$ & $0.8265\pm 0.1585$ \\
General & $\vec{c}_{n,t}^\top\vec{W} \vec{u}_m$ & $0.7802\pm 0.1439$ \\ \bottomrule
\end{tabular}
\label{table:attention_results}
\end{table*}

\subsection{Number of Experts}

\begin{figure}[htbp]
\centering
\includegraphics[width=0.5\linewidth]{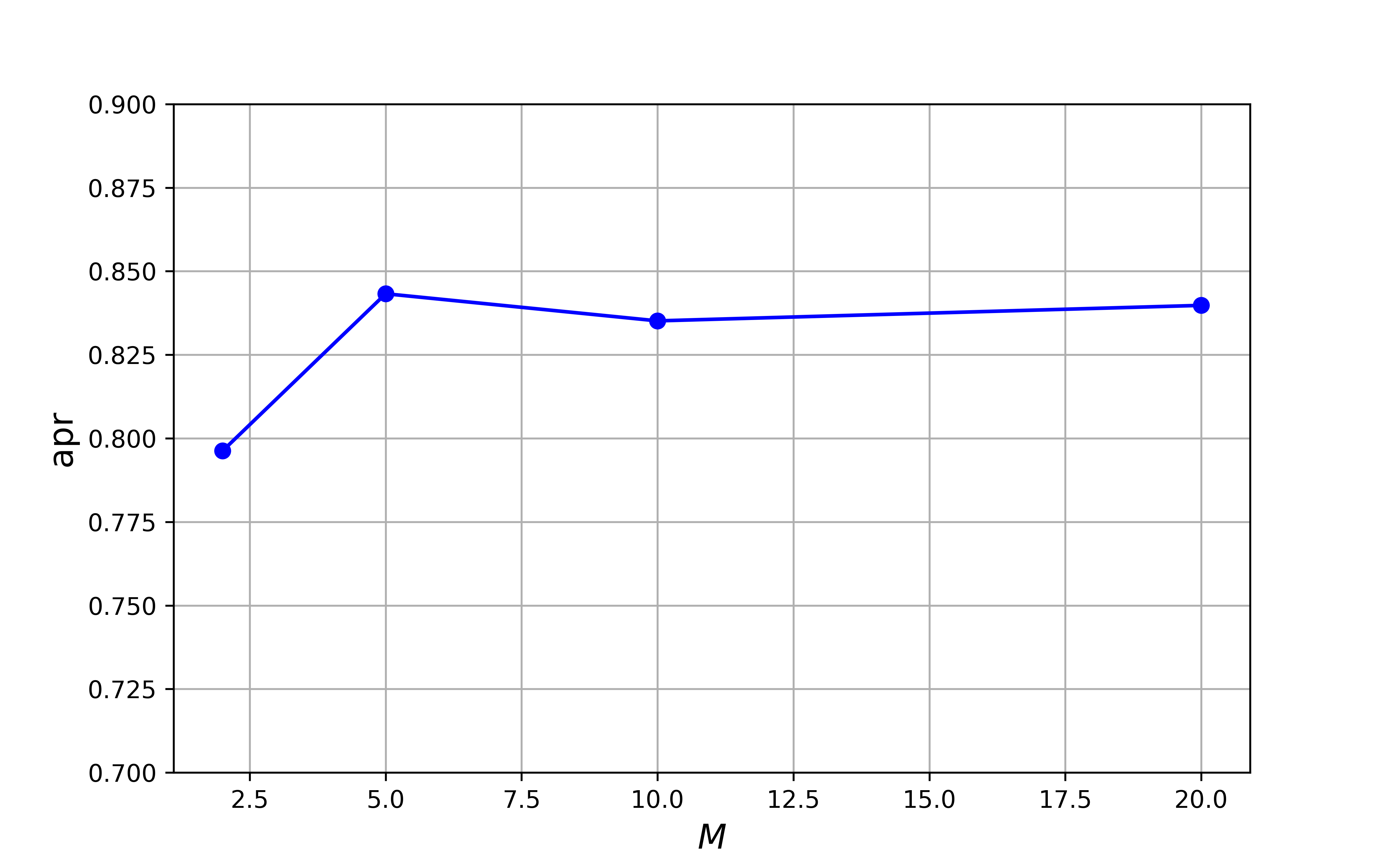}
\caption{Validation performance of MAES on the $\delta=0.2$ dataset, with increasing numbers of experts $M$. Increasing the number of experts initially improves performance, as these experts can specialize on different distributions of the data. However, the performance eventually plateaus and begins decreasing due to the soft nature of the attention gate, which assigns non-zero weights to under-performing experts, thereby impeding on the training of the better-performing ones as more unnecessary experts are added to the ensemble.}
\label{fig:adaptive_nexp}
\end{figure}

We then varied the number of experts $M$, each randomly sampled from the $20$ LSTMs (Section 5.1.2 in the main paper). From Figure \ref{fig:adaptive_nexp}, it can be observed that increasing the number of experts generally improved performance. By including more experts, we are more likely to find an architecture that performs better on a given subset of the data, and the gate can simply assign a negligible weight to under-performing experts.

Nonetheless, the soft attention mechanism still assigns non-zero weights to all experts. With more distributed attention weights, the better-performing experts are assigned smaller weights with larger $M$, thus resulting in ``weaker'' updates for the latter and thereby affecting the ensemble performance (Figure \ref{fig:adaptive_nexp} for $M>5$). Soft attention also slows down training for larger $M$, as all the experts' parameters are updated for each step of gradient descent. Clipping infinitesimal weights could accelerate the training procedure, but might cause some experts to be disregarded prematurely.

The best-performing MAES architecture consisted of $M=5$ experts and an Additive attention mechanism for the gate, trained jointly with the bespoke loss function $\mathcal{L}$. This configuration was maintained when evaluating MAES on the test datasets for all amounts of temporal conditional shift (Section 5.2).

\bibliography{MAES}

\vfill